\documentclass[lettersize,journal]{IEEEtran}
\usepackage{amsmath,amsfonts}
\usepackage{algorithmic}
\usepackage{algorithm}
\usepackage{array}
\usepackage[caption=false,font=normalsize,labelfont=sf,textfont=sf]{subfig}
\usepackage{amsthm}

\usepackage{textcomp}
\usepackage{stfloats}
\usepackage{url}
\usepackage{verbatim}
\usepackage{graphicx}
\usepackage{multirow}
\usepackage{booktabs}
\usepackage{amssymb}
\def\BibTeX{{\rm B\kern-.05em{\sc i\kern-.025em b}\kern-.08em
    T\kern-.1667em\lower.7ex\hbox{E}\kern-.125emX}}
\usepackage{balance}
\usepackage[table]{xcolor} 
\usepackage{hyperref}
\hypersetup{
    colorlinks=true,
    linkcolor=blue,
    filecolor=magenta,      
    urlcolor=cyan,
    pdftitle={Spatial Attention: Adapting Execution Horizons for Diffusion Policies via Observation Sensitivity},
    pdfauthor={Che-Sang Park, Junsu Ha, Jianlong Fu, Frank C. Park},
    pdfpagemode=FullScreen,
    }

\begin{document}
\title{Spatial Attention: Adapting Execution Horizons \\ for Diffusion Policies via Observation Sensitivity}

\author{Che-Sang Park, Junsu Ha, Jianlong Fu, and Frank C. Park%
\thanks{Under Review.}%
\thanks{Che-Sang Park, Junsu Ha, and Frank C. Park are with the Robotics Laboratory, Seoul National University, Seoul, South Korea (e-mail: \{cspark, hajunsu\}@robotics.snu.ac.kr;
fcp@snu.ac.kr).}%
\thanks{Jianlong Fu is with Microsoft Research Asia, Beijing, China
(e-mail: jianf@microsoft.com).}}

\markboth{IEEE ROBOTICS AND AUTOMATION LETTERS,~Vol.~11, No.~6, June~2026}%
{Shell \MakeLowercase{\textit{et al.}}: A Sample Article Using IEEEtran.cls for IEEE Journals}

\maketitle

\begin{abstract}

Sampling action chunks via generative models has become a widely adopted methodology for robotic learning from demonstration.
However, existing methods often struggle to balance responsiveness and computational cost because they execute each action chunk for a fixed execution horizon.
In this paper, we adaptively adjust the execution horizon of sampled action chunks, balancing responsiveness and computational efficiency.
We introduce Spatial Attention---defined as the expected squared norm of the gradient of the action log-likelihood with respect to the observation---which indicates the sensitivity of the policy's action distribution to variations in the observation.
We show that, under a fixed budget of chunk samplings, the execution horizon that minimizes the cumulative likelihood drop induced by disturbances decreases as Spatial Attention increases.
By forecasting future Spatial Attention values alongside the action chunk, our framework dynamically assigns shorter execution horizons to phases with high Spatial Attention, and longer horizons to phases with low Spatial Attention.
Experiments on standard and perturbed tasks, in both simulation and on a real robot, show that our method significantly improves success rates over fixed-horizon baselines while maintaining the average execution horizon.

\end{abstract}


\section{Introduction}
\label{sec:intro}

Consider a robotic system with state space $\mathcal{X}$, observation space $\mathcal{O}$, and action space $\mathcal{A}$, governed by $x_{k+1} = f(x_k, a_k) + n_k$ and $o_k = h(x_k) + w_k$, where $x_k \in \mathcal{X}$, $o_k \in \mathcal{O}$, and $a_k \in \mathcal{A}$ denote the state, observation, and action at timestep $k$, and $n_k$ and $w_k$ are process and observation noise.
The dynamics $f$ and the observation function $h$ vary with the embodiment and sensing modality.
Since both $f$ and $h$ are unknown, learning from demonstration trains a policy $\pi(a | o)$ that infers actions directly from observations, using a demonstration dataset $\mathcal{D} = \{(o_i, a_i)\}_{i=1, 2\cdots}$.
Recent robotic policies built on diffusion models~\cite{chi2023diffusion, prasad2024consistency, wang2024one} model $\pi$ as a conditional generative model over action sequences.
Given an observation $o_k$, the policy samples a sequence of actions $[a_k, a_{k+1}, \ldots, a_{k+T_H-1}] \sim \pi(\cdot | o_k)$, referred to as an \textit{action chunk}~\cite{zhao2023learning}.
The policy then executes only the first $T_a$ ($T_a \leq T_H$) actions of the chunk before sampling a new one, in a receding-horizon fashion similar to model predictive control; this $T_a$, referred to as the \textit{execution horizon}, is fixed to a user-specified value prior to deployment.

Executing an action chunk, however, leaves the policy vulnerable to external disturbances and errors, since a new observation is incorporated only after all $T_a$ actions have been executed~\cite{chen2025responsive, xue2025reactive}.
Reducing the execution horizon allows more frequent observations and thus improves responsiveness, but each additional chunk sampling invokes the slow generative process, increasing computational cost and overall task completion time~\cite{prasad2024consistency, wang2024one, xie2024subconscious, kim2024openvla, black2025real}.
A fixed execution horizon therefore forces a single compromise between responsiveness and efficiency for the entire task, even though different phases of a task demand different levels of responsiveness.

This motivates adapting the execution horizon during deployment, rather than fixing it in advance.
Recently, several works in the domain of planning with diffusion models have employed the concept of adaptive replanning~\cite{zhou2023adaptive, jutras2024adaptive, punyamoorty2024dynamic}.
However, directly applying these methods to a generative policy inherently requires the generation of future observations to determine the optimal moment to replan.
This computational bottleneck makes such approaches impractical for tasks with high-dimensional observations~\cite{chi2023diffusion}.

In this work, we propose \emph{Spatial Attention}, a criteria for adaptively determining the execution horizon of a sampled action chunk.
Motivated by minimum-attention control~\cite{brockett1997minimum, jang2015minimum}, which measures a control law's attention through its sensitivity to changes in state and time, we adopt the spatial (state) component and recast it as the sensitivity of the policy's action distribution to its observation.
By formulating an optimization problem that minimizes the cumulative likelihood drop of sampled actions under a fixed average sampling frequency, we show that the optimal execution horizon decreases as Spatial Attention---the expected squared norm of the gradient of the action log-likelihood with respect to the observation---increases.
At inference time, a sequence-to-sequence estimator forecasts future Spatial Attention from the current observation and the sampled action chunk, and the policy sets the execution horizon to the point at which the accumulated prediction reaches a predefined threshold---assigning shorter horizons in high-attention regions and longer ones in low-attention regions.
Because Spatial Attention is estimated from auxiliary score networks trained on the demonstration data, independently of the policy's architecture, our framework applies to any policy that generates action chunks.

We extensively evaluate our approach on robotic manipulation tasks across both simulation and real-world environments. 
In simulation, we utilize the standard Robomimic benchmark~\cite{robomimic2021} to verify general performance, and further introduce manual perturbations during execution to rigorously evaluate the policy's responsiveness to external disturbances. 
We also conduct real-world experiments where dynamic disturbances are applied to the target object to assess the real-time capability of our approach. 
Experimental results demonstrate that dynamically adjusting the execution horizon based on Spatial Attention improves success rates across all tasks while maintaining the average sampling frequency, confirming that our approach enhances responsiveness without additional computational overhead.

\section{Related Works}
\label{sec:related_works}

\begin{figure*}
    \centering
    \vspace*{3mm}
    \includegraphics[width=0.95\linewidth]{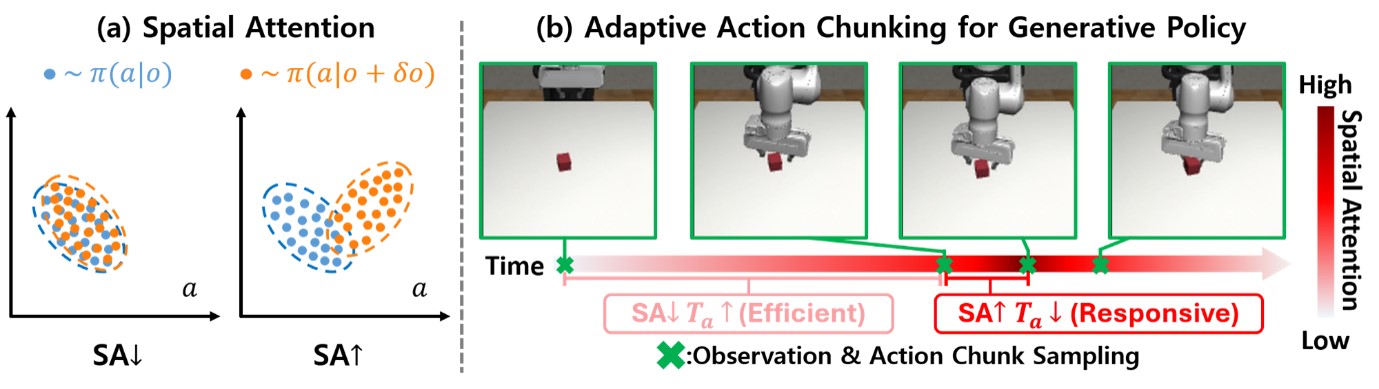}
    \caption{(a) Spatial Attention (SA) quantifies the sensitivity of the policy's probability density function to variations in the observation. While a distribution with low SA remains relatively stable under observation perturbations, a high-SA distribution shifts significantly to even minor perturbations. (b) In this work, we utilize SA to adaptively adjust the execution horizon, $T_a$. Extending $T_a$ during low-SA phases improves computational efficiency, while shortening $T_a$ in high-SA regions ensures robustness and responsiveness to errors or environmental changes.}
    \vspace*{-3mm}
    \label{fig:Overall Concept}
\end{figure*}

\subsection{Action Chunking in Generative Policies}

Recent policies built upon generative models commonly sample a sequence of actions from a single observation, referred to as an action chunk, rather than predicting a single-step action~\cite{zhao2023learning}.
Compared to single-step action prediction, action chunking offers several practical advantages.
It mitigates compounding errors in long-horizon tasks by reducing the number of autoregressive prediction steps~\cite{zhao2023learning, zhang2025action}.
Moreover, it maintains temporal consistency by capturing non-Markovian behaviors such as pauses and oscillating motions~\cite{chi2023diffusion, chen2025responsive, xue2025reactive}.
Beyond these benefits, it also enables computationally expensive Vision-Language-Action (VLA) models to control physical robots by reducing the frequency of queries to the VLM backbone~\cite{kim2024openvla, black2024pi_0, gr00tn1_2025}.
Despite these benefits, executing action chunks lacks the ability to balance computational efficiency against responsiveness to external disturbances, since the execution horizon is fixed.

\subsection{Adaptive Execution Horizons in Generative Planners and Policies }

Adaptively adjusting the execution horizon has been explored in both planning and policies that use generative models.
In the field of planning with generative models~\cite{janner2022planning, ajay2022conditional}, several works have investigated when to trigger replanning~\cite{zhou2023adaptive, jutras2024adaptive, punyamoorty2024dynamic}.
RDM~\cite{zhou2023adaptive} uses the likelihood of the planned trajectory as the replanning criterion.
After initial planning with Diffuser~\cite{janner2022planning}, RDM monitors the trajectory likelihood at each timestep and triggers replanning when it falls below a predefined threshold.
Other works, such as Jutras-Dub\'e et al.~\cite{jutras2024adaptive} and Punyamoorty et al.~\cite{punyamoorty2024dynamic}, instead use predictive uncertainty as the replanning criterion.
These methods first generate a sequence of future states using Decision Diffuser~\cite{ajay2022conditional}, and then infer an action at each timestep from two consecutive states using an ensemble of inverse dynamics models.
A predictive uncertainty estimate, computed from the ensemble's action predictions, then serves as the signal that determines whether to replan.
Despite the benefits of replanning, these methods require generating future states, which is impractical for real-time control with high-dimensional visual observations.

Adaptive execution-horizon adjustment has also been studied in the context of generative policies, which avoid this limitation~\cite{jing2025mixture}.
Unlike planners, generative policies directly produce actions without generating future observations, making them better suited to real-time control with high-dimensional visual inputs.
Mixture of Horizons (MoH)~\cite{jing2025mixture} produces action predictions at multiple horizons together with a fused prediction.
At inference, MoH measures the disagreement between the fused prediction and each horizon-wise prediction at every step within the chunk, and executes the longest prefix whose disagreement remains below a threshold derived from the early steps.
Despite this effort, the proposed metric is heuristic and lacks mathematical proof of optimality.
Furthermore, this work does not consider scenarios with external disturbances.
In this work, we establish a mathematically grounded metric for adaptive execution horizon derived from an optimization problem.
We further evaluate our method under disturbance scenarios to demonstrate its robustness.

\subsection{Score-Based Generative Models}
We briefly review score-based generative models, as they form the mathematical basis for our derivation of Spatial Attention.
Score-based generative models generate samples from noise through a denoising process known as Annealed Stochastic Langevin Dynamics~\cite{song2019generative, song2020improved}.
This discrete sampling process can be interpreted as solving a reverse-time Stochastic Differential Equation (SDE)~\cite{song2020score}.
While the forward SDE defines a fixed process that gradually adds noise to the data, the generative process requires reversing these dynamics.
Unlike the forward SDE, where the dynamics are given, the reverse process depends on the score of the target distribution, which must be estimated.
Consequently, the primary objective of these models is to learn a score function $s_\theta(x, t)$ that approximates the gradient of the log-density of perturbed data at a given noise scale.
\begin{equation}
    s_\theta(x, t) = \nabla_x \log p_t(x).
\end{equation}
Depending on the specific formulation of the forward dynamics and the training objective, this framework gives rise to several model classes, such as Denoising Score Matching (DSM)~\cite{vincent2011connection}, Noise Conditional Score Networks (NCSN)~\cite{song2019generative, song2020improved}, and Denoising Diffusion Probabilistic Models (DDPM)~\cite{ho2020denoising, sohl2015deep}.
We leverage this score estimation framework, specifically NCSN~\cite{song2019generative}, to derive Spatial Attention in Section~\ref{sec:adp}.

\section{Spatial Attention: Maximizing Policy Robustness with Fixed Sampling Budget} \label{sec:max_robustness}

In this section, we begin by formulating how external disturbances degrade the policy's action log-likelihood when executing a long action chunk, and we introduce the strategy used to mitigate this degradation.
We then formulate the problem of minimizing the cumulative likelihood drop subject to a constraint on the total number of resampling events over a task horizon.
Solving this problem reveals that policy robustness is maximized when the resampling frequency scales with the gradient of the log-likelihood of the policy with respect to its observation --- a quantity we refer to as \textit{Spatial Attention}.

\subsection{Likelihood Drop Under Action Chunk Execution}
Given an observation $o\in \mathbb{R}^{dim(o)}$, generative policies sample an action chunk of length $T_H$.
An action chunk can be seen as a continuous function $\mathbf{A}: [0, T_H] \to \mathcal{A}$, where $\mathcal{A}$ is the action space. Under this convention, the policy samples
\begin{equation}
    \mathbf{A} \sim \pi(\mathbf{A} \mid o),
\end{equation}
where $\pi$ is the true distribution that the policy approximates.

Ideally, if no external disturbance is given during rollout and the policy is well-trained, one could set the execution horizon $T_a$ ($T_a \leq T_H$) equal to the full chunk length $T_H$ from the initial observation to complete the task with minimal resampling, maximizing computational efficiency.
However, this strategy is extremely vulnerable to external disturbances or errors since the policy receives no updated observation during chunk execution.
Specifically, if the observation is perturbed by a disturbance $\delta o$, the expected log-likelihood drop of action $a\in \mathcal{A}$ is:
\begin{equation}
    \mathbb{E}_{a \sim \pi(\cdot \mid o)} \left[ \log \pi(a \mid o) - \log \pi(a \mid o + \delta o) \right].
\end{equation}

In general, as action chunk execution continues without incorporating new observations, disturbances accumulate, increasing the likelihood drop and degrading policy performance. We model the accumulated disturbance $\delta o$ at elapsed time $\tau$ since the last observation as zero-mean Gaussian noise whose variance grows as a power of $\tau$:
\begin{equation} \label{eq:disturbance}
\delta o \sim \mathcal{N}\!\left(0,\, \epsilon^2\, \tau^{2\gamma}\, \mathbb{I}\right),
\end{equation}
where $\epsilon > 0$ is a noise scale and $\gamma > 0$ is a hyperparameter controlling how rapidly the disturbance grows with the open-loop execution time.

To balance robustness and computational efficiency, we introduce a time-dependent resampling rate function $\rho(t)$. Whenever a new chunk is sampled at time $t_k$, its execution horizon $T_a$ is chosen such that the integral of $\rho(t)$ over the chunk satisfies:
\begin{equation} \label{eq:chunk_constraint}
    \int_{t_k}^{t_k + T_a} \rho(t)\, dt = 1.
\end{equation}

Letting $T_{\max}$ denote the task horizon, the total number of resampling events over the rollout is $N = \int_0^{T_{\max}} \rho(t)\, dt$. The optimal $\rho(t)$ that minimizes the expected likelihood drop subject to a fixed sampling budget $N$ is derived in Section~\ref{subsec:obs_opt}, and its closed-form solution is given in Section~\ref{subsec:spatial_attention_1}.

\subsection{Minimizing Likelihood Drop under a Fixed Budget} \label{subsec:obs_opt}

Let
\begin{equation} \label{eq:drop_def}
    \mathcal{D}(o, \tau) \triangleq \mathbb{E}_{\delta o,\, a}\!\left[ \log \pi(a|o) - \log \pi(a|o+\delta o) \right],
\end{equation}
denote the expected log-likelihood drop after the policy has executed open-loop for an elapsed time $\tau$, where the variance of the disturbance $\delta o$ follows Equation~\ref{eq:disturbance}.

We seek the resampling rate $\rho(t)$ that minimizes the cumulative expected drop under a fixed $N$ sampling budget. Under rate $\rho(\cdot)$, the policy resamples about $\rho(t)$ times per unit time and runs open-loop for roughly $1/\rho(t)$ before each resampling. Accumulating $\mathcal{D}$ over every open-loop interval gives the ideal objective:
\begin{align}
\begin{split}
\label{eq:ideal_objective}
\arg\min_{\rho(\cdot)} \quad & \int_{0}^{T_{\max}} \rho(t) \int_{0}^{1/\rho(t)} \mathcal{D}(o(t), \tau)\, d\tau\, dt \\
\text{s.t.} \quad & \int_{0}^{T_{\max}} \rho(t)\, dt = N.
\end{split}
\end{align}

Evaluating Equation~\ref{eq:ideal_objective} directly is impractical: $\pi$ does not provide an explicit probability density, so $\mathcal{D}$ has no closed form and cannot be optimized during rollout.
We therefore reduce it to the tractable problem
\begin{align}
\begin{split}
\label{eq:minimize_by_rho_}
\arg\min_{\rho(\cdot)} \quad & \int_{0}^{T_{\max}} \mathrm{Att}(o(t))\, \left(\frac{1}{\rho(t)}\right)^{2\gamma} dt \\
\text{s.t.} \quad & \int_{0}^{T_{\max}} \rho(t)\, dt = N,
\end{split}
\end{align}
where $\mathrm{Att}(o) \triangleq \mathbb{E}_{a \sim \pi(\cdot|o)}\!\left[\| \nabla_o \log \pi(a|o) \|^2\right]$ measures the policy's sensitivity to observation perturbations. The rest of this subsection derives Equation~\ref{eq:minimize_by_rho_} from Equation~\ref{eq:ideal_objective} by approximating $\mathcal{D}$ in closed form.

A second-order Taylor expansion of $\log \pi(a|o+\delta o)$ around $o$ gives
\begin{equation}
\begin{aligned}
    \log \pi(a | o) & - \log \pi(a | o + \delta o) \\
    \approx{}  -\nabla_o \log \pi(a | o)^\top \delta o & - \tfrac{1}{2}\, \delta o^\top \nabla_o^2 \log \pi(a | o)\, \delta o.
\end{aligned}
\end{equation}

Taking expectation over $\delta o$ removes the linear term, yielding
\begin{equation}
    \mathcal{D}(o,\tau) \approx -\frac{\epsilon^2\tau^{2\gamma}}{2}\, \mathbb{E}_{a}\!\left[ \mathrm{tr}\, \nabla_o^2 \log \pi(a | o) \right].
\end{equation}
Applying the Fisher-information identity
\begin{equation*}
    \mathbb{E}_a\!\left[\| \nabla_o \log \pi(a | o) \|^2\right] = -\mathbb{E}_a\!\left[\mathrm{tr}\, \nabla_o^2 \log \pi(a | o)\right],
\end{equation*}
the per-step drop at elapsed time $\tau$ becomes
\begin{equation} \label{eq:per_step_drop}
    \mathcal{D}(o,\tau) \approx \tfrac{1}{2}\, \mathrm{Att}(o)\, \epsilon^2\, \tau^{2\gamma}.
\end{equation}

Now consider a sub-interval $[k\Delta T,\, (k+1)\Delta T]$ on which $\rho(\cdot)$ takes minimum and maximum values $\rho_{\min}, \rho_{\max}$, and on which $\mathrm{Att}(o(\cdot))$ takes extrema $\mathrm{Att}_{\min}, \mathrm{Att}_{\max}$.
Within this sub-interval, the policy resamples between $\rho_{\min}\Delta T$ and $\rho_{\max}\Delta T$ times, and each chunk has duration between $1/\rho_{\max}$ and $1/\rho_{\min}$.
Integrating Equation~\ref{eq:per_step_drop} over a single chunk and summing over chunks, the cumulative expected drop $Err$ on the sub-interval is bounded as
\begin{equation} \label{eq:inequality_}
    \tfrac{\epsilon^2\,\Delta T}{2(2\gamma+1)}\,\mathrm{Att}_{\min}\,\rho_{\max}^{-2\gamma} \leq Err \leq \tfrac{\epsilon^2\,\Delta T}{2(2\gamma+1)}\,\mathrm{Att}_{\max}\,\rho_{\min}^{-2\gamma}.
\end{equation}
Assuming $\mathrm{Att}(o(\cdot))$ and $\rho(\cdot)$ are continuous, taking $\Delta T \to 0$ yields
\begin{equation}
    \frac{dErr}{dt} \;\to\; \frac{\epsilon^2}{2(2\gamma+1)}\, \mathrm{Att}(o(t))\,\rho(t)^{-2\gamma}.
\end{equation}
Integrating over $[0, T_{\max}]$ and absorbing the $\rho$-independent constant $\epsilon^2/[2(2\gamma+1)]$ recovers the tractable objective in Equation~\ref{eq:minimize_by_rho_}.

\subsection{Spatial Attention} \label{subsec:spatial_attention_1}
The solution of the constrained optimization in Equation~\ref{eq:minimize_by_rho_} can be derived via a standard Lagrangian argument. Pointwise stationarity (justified by convexity of the integrand in $\rho$) gives $\rho(t)^{2\gamma+1} \propto \mathrm{Att}(o(t))$, equivalently
\begin{equation} \label{eq:spatial attention}
    \rho(t) \;\propto\; \mathrm{Att}(o(t))^{\,1/(2\gamma+1)},
\end{equation}
where the proportionality constant is fixed by the budget constraint in Equation~\ref{eq:minimize_by_rho_}.

We refer to the quantity
\begin{equation} \label{eq:spatial_attention_def}
    \mathrm{Att}(o) \;\triangleq\; \mathbb{E}_{a \sim \pi(\cdot | o)}\!\left[\, \big\|\nabla_o \log \pi(a | o) \big\|^{2}\,\right]
\end{equation}
as \textit{Spatial Attention}. 
It quantifies the squared sensitivity of the policy's action distribution to perturbations in the observation: when Spatial Attention is large, even small observation changes induce a large shift in the action distribution, and the optimal strategy is to resample more frequently (i.e., shorten the execution horizon $T_a$); when Spatial Attention is small, longer chunks suffice. 
The execution horizon $T_a$ for each chunk is determined by Equation~\ref{eq:chunk_constraint} applied to the optimal $\rho(t)$ from Equation~\ref{eq:spatial attention}.

\section{Spatial Attention as a Module for Adaptive Execution Horizon}
\label{sec:adp}

In this section, we introduce practical challenges when deriving Spatial Attention and methods to resolve these issues.
Along with it, we propose our adaptive execution horizon selection pipeline using Spatial Attention.
Section~\ref{subsec:conditiona_gradient_estimation} shows how to obtain Spatial Attention when the policy does not provide an explicit probability density function.
Then Section~\ref{subsec:vision_based_cge} extends this method to the case where the observation is a vision input.
Finally, we present our full inference-time pipeline driven by Spatial Attention in Section~\ref{subsec:forecasting}.

\subsection{Deriving Spatial Attention with Bayes' Rule}\label{subsec:conditiona_gradient_estimation}

In this section, we present the key challenge in deriving \textit{Spatial Attention} from Equation~\ref{eq:spatial attention} and our approach to overcoming it.
The main bottleneck in computing \textit{Spatial Attention} is evaluating
\begin{equation}\label{eq:cge}
    \nabla_o\log\pi(a|o).
\end{equation}
This is because $\pi(a|o)$ is a generative model's distribution that does not provide an explicit probability density function.
Moreover, even when the distribution is a score-based generative model, we cannot directly reuse its score estimate $s_\theta(a, t) \approx \nabla_a \log \pi(a|o)$, since Equation~\ref{eq:cge} is a gradient with respect to the observation $o$, not the action $a$.

We overcome this limitation by following the Bayes' rule decomposition used in classifier guidance~\cite{dhariwal2021diffusion}.
Specifically, we reformulate Equation~\ref{eq:cge} as:
\begin{equation}\label{eq:cge_2}
    \nabla_o \log \pi(a|o) = \nabla_o \log \pi(o|a) - \nabla_o \log \pi(o),
\end{equation}
which follows from $\log \pi(a|o) = \log \pi(o|a) + \log \pi(a) - \log \pi(o)$ and $\nabla_o \log \pi(a) = 0$.
The two terms on the right-hand side are the score of the action-conditional observation distribution $\pi(o|a)$ and the score of the marginal observation distribution $\pi(o)$, respectively.
We therefore train two separate score-based generative models, one for each distribution, and use their score estimates to evaluate Equation~\ref{eq:cge}.
In this work, we adopt NCSN~\cite{song2019generative} as our score-based generative model.

\subsection{Spatial Attention from Vision Input}\label{subsec:vision_based_cge}
In the case of using image observation, several challenges arise when deriving Spatial Attention.
Due to the high dimensionality of image observations, training a separate score-based generative model for deriving Spatial Attention requires high computation cost~\cite{rombach2022high}.
Moreover, taking the gradient directly with respect to the image observation measures the change of the policy distribution with respect to pixel-level perturbations, which is not ideal since many pixels are irrelevant to the task.

We instead evaluate Spatial Attention in the latent space of a VAE~\cite{kingma2013auto}, which we treat as a practical proxy for observation-space sensitivity:
\begin{equation} \label{eq:latent_spatial_attention}
    \mathrm{Att}(z) \triangleq \mathbb{E}_{a\sim\pi(a|z)}\left[\left\|\nabla_z \log \pi(a|z)\right\|^2\right], \quad z = f_\psi(o).
\end{equation}
Since the encoder is trained to capture task-relevant structure of the observation, the latent-space gradient norm reflects the policy's sensitivity to meaningful observation changes while excluding pixel-level noise. The encoder $f_\psi$ is trained as part of the VAE on the training data.

\subsection{Adaptive Execution Horizon via Forecasted Spatial Attention}
\label{subsec:forecasting}
We now describe how our framework adaptively selects the execution horizon $T_a$ of the sampled action chunk $\mathbf{A}_t$ of length $T_H$.
To obtain future Spatial Attention values without generating future observations, we employ a sequence-to-sequence model $\mathbf{Att}_\phi$ that forecasts the Spatial Attention sequence from the current observation and the sampled action chunk:
\begin{equation}
    [\mathrm{Att}_t, \mathrm{Att}_{t+1}, \ldots, \mathrm{Att}_{t+T_H-1}] = \mathbf{Att}_\phi(\mathbf{A}_t, o_t),
\end{equation}
where $\mathrm{Att}_t$ denotes the Spatial Attention at time $t$, equal to $\mathrm{Att}(o_t)$ and $\mathrm{Att}(z_t)$ for state and vision observations, respectively.

From Equation~\ref{eq:spatial attention} and Equation~\ref{eq:chunk_constraint}, $T_a$ is determined by integrating the optimal resampling rate over the chunk.
Discretizing this integral over the forecasted Spatial Attention sequence yields:
\begin{equation}
    T_a = \min_{k} \Big\{ k \,\Big|\, \sum_{i=0}^{k}{\mathrm{Att}_{t+i}^{1/(2\gamma+1)}} > C_{\mathrm{att}} \Big\}.
\label{eq:cumulative}
\end{equation}
$C_{\mathrm{att}}$ is a hyperparameter fixed prior to rollout.
If the Spatial Attention values are too small for Equation~\ref{eq:cumulative} to have a solution, the full action chunk is executed (i.e., $T_a = T_H$).

The forecasting model $\mathbf{Att}_\phi$ is trained offline on a dataset constructed from the policy's training trajectories, where each trajectory is labeled with the Spatial Attention values computed using the score networks from Section~\ref{subsec:conditiona_gradient_estimation}.
We use a transformer architecture~\cite{vaswani2017attention} for $\mathbf{Att}_\phi$.

\section{Experiment}\label{sec:experiment}

In this section, we evaluate applying Spatial Attention to execution horizon decisions on a standard simulation benchmark, Robomimic~\cite{robomimic2021}, to demonstrate its general performance.
We then evaluate on a modified simulation environment in which external disturbances are applied during rollouts to show its responsiveness to disturbances.
Lastly, we deploy Spatial Attention on a real-world robot task to demonstrate its real-time effectiveness on a physical robot.
The experimental results show that using Spatial Attention as a criterion for the execution horizon consistently improves the success rate under the same average execution horizon.

\begin{figure*}
    \centering
    \includegraphics[width=0.95\linewidth]{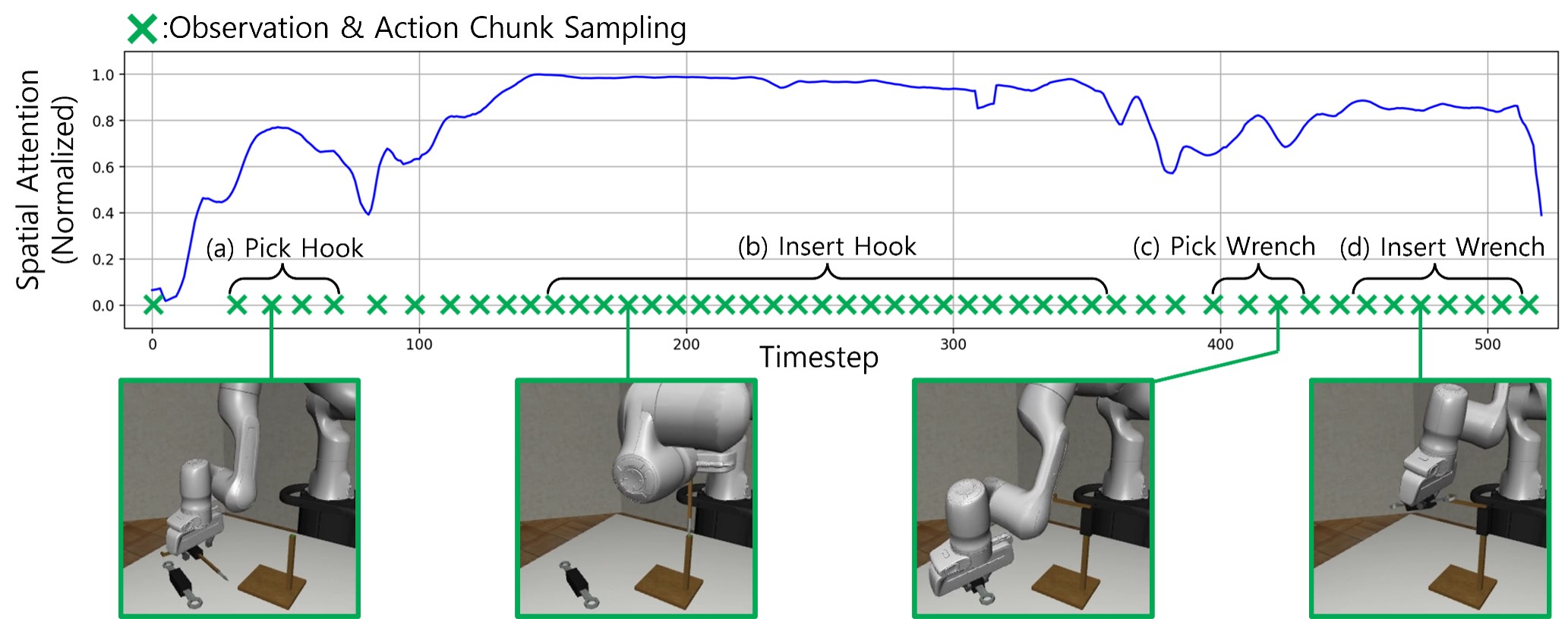}
    \vspace*{-3mm}
    \caption{\textbf{Spatial Attention values during Tool Hang task.}
    The top plot illustrates the evolution of Spatial Attention throughout the task, where green crosses mark the moments of observation and action chunk sampling.
    The frames below depict the physical states of the robot corresponding to these connected crosses.
    Notably, Spatial Attention increases during important manipulation phases, such as grasping ((a), (c)) and inserting ((b), (d)) the target objects.
    Our adaptive execution horizon pipeline utilizes this metric to dynamically shorten the execution horizon, $T_a$, in these critical regions.
    Consequently, the robot becomes more responsive to potential errors and disturbances during these moments, all while preserving computational cost by maintaining a consistent average execution horizon.
    }
    \label{fig:tool_hang}
\end{figure*}

\subsection{Baselines}
We evaluate our method, denoted +SA, by applying Spatial Attention on top of each of the following three baseline policies and comparing each against its fixed-length baseline. 
These baselines execute each sampled action chunk for a fixed execution horizon $T_a$, set as a hyperparameter and held constant throughout the rollout.

\subsubsection{Diffusion Policy (DDPM / DDIM)~\cite{chi2023diffusion}}

As one of the first methods to adapt diffusion models to learning from demonstration, DP samples an action chunk via a diffusion process from a given observation.
The type of diffusion process determines the number of diffusion steps required to sample a single action chunk.
In this experiment, we use both DDPM~\cite{ho2020denoising} and DDIM~\cite{song2020denoising} as baselines, for which the total diffusion steps needed for sampling are 100 and 16, respectively.
The UNet architecture takes two observation frames as input and estimates the score required for sampling.

\subsubsection{Consistency Policy (CP)~\cite{prasad2024consistency}}
To reduce the computation cost of DDPM \& DDIM even more, Consistency Policy~\cite{prasad2024consistency} 
brings Consistency Trajectory Models (CTM)~\cite{kim2023consistency} into Diffusion Policy framework.
Taking advantage of CTM, the sampler only needs 3 diffusion steps to sample high-quality data and 1 step to sample marginal-quality data.
In this experiment, we sample action chunk with both 3-step CP and 1-step CP for evaluation.
As noted in~\cite{prasad2024consistency}, the architecture, infrastructure, and input/output formats are identical to those of Diffusion Policy.

\subsection{Evaluation Methodology}
Policies are trained with 4 different training seeds. 
For each seed, we save the best 5 checkpoints, selected based on evaluations performed every 50 epochs.  
This results in a total of 20 checkpoints, over which we compute the average and standard deviation of the performance.

We evaluate both the baseline and our proposed algorithms using the average success rate across 100 evaluation episodes for each task.
For the adaptive execution horizon with Spatial Attention, we select an appropriate $C_{att}$ such that the resulting mean execution horizon, $T_a$, for each episode falls within the range $[T_{avg}-0.5, T_{avg}+0.5]$, where $T_{avg}$ represents the fixed $T_a$ value used for DDPM/DDIM and CP.
This setting ensures a fair comparison, as execution horizon significantly impacts policy performance and its computational efficiency.

\begin{table*}
\centering
\vspace{3mm}
{
\setlength{\aboverulesep}{0pt}
\setlength{\belowrulesep}{0pt}
\renewcommand{\arraystretch}{1.2}

\caption{
    \textbf{Simulation Evaluation Results.} 
    We report the success rates of our method (denoted +SA) and its corresponding baselines evaluated on the Robomimic~\cite{robomimic2021} benchmark. 
    Success rates are measured by evaluating a single checkpoint under 100 different initial conditions. 
    The average and standard deviation of these success rates are derived by evaluating the 5 best checkpoints during training across 4 different seeds, totaling 20 checkpoints. 
}
\begin{tabular}{@{}l|c|cccc|cccc@{}}
\toprule

\multirow{3}{*}{Method} & \multirow{3}{*}{Steps} & \multicolumn{4}{c|}{ Robomimic Simple ($T_{avg}=16$)} & \multicolumn{4}{c}{Robomimic Complex ($T_{avg}=32$)} \\ 
\cmidrule(lr){3-6} \cmidrule(lr){7-10} 

 & & \multicolumn{2}{c}{Lift} & \multicolumn{2}{c|}{Can} & \multicolumn{2}{c}{Square} & \multicolumn{2}{c}{Tool Hang} \\ 
\cmidrule(lr){3-4} \cmidrule(lr){5-6} \cmidrule(lr){7-8} \cmidrule(lr){9-10}

 & & State & Vision & State & Vision & State & Vision & State & Vision \\
 
\noalign{\global\aboverulesep=0.4ex \global\belowrulesep=0.65ex}

\midrule

DDPM & \multirow{2}{*}{100} & 0.98{\scriptsize{$\pm$0.010}} & \textbf{0.99}{\scriptsize{$\pm$\textbf{0.004}}} & 0.96{\scriptsize{$\pm$0.009}} & \textbf{0.98}{\scriptsize{$\pm$\textbf{0.012}}} & 0.89{\scriptsize{$\pm$0.022}} & 0.75{\scriptsize{$\pm$0.041}} & 0.47{\scriptsize{$\pm$0.040}} & 0.61{\scriptsize{$\pm$0.052}} \\
DDPM+SA (ours) & & \textbf{0.99}{\scriptsize{$\pm$\textbf{0.008}}} & \textbf{0.99}{\scriptsize{$\pm$\textbf{0.004}}} & \textbf{0.98}{\scriptsize{$\pm$\textbf{0.009}}} & \textbf{0.98}{\scriptsize{$\pm$\textbf{0.011}}} & \textbf{0.94}{\scriptsize{$\pm$\textbf{0.022}}} & \textbf{0.80}{\scriptsize{$\pm$\textbf{0.043}}} & \textbf{0.57}{\scriptsize{$\pm$\textbf{0.072}}} & \textbf{0.75}{\scriptsize{$\pm$\textbf{0.047}}} \\
\midrule

DDIM & \multirow{2}{*}{16} & \textbf{0.99}{\scriptsize{$\pm$\textbf{0.009}}} & \textbf{0.99}{\scriptsize{$\pm$\textbf{0.004}}} & 0.95{\scriptsize{$\pm$0.011}} & 0.98{\scriptsize{$\pm$0.011}} & 0.87{\scriptsize{$\pm$0.030}} & 0.80{\scriptsize{$\pm$0.021}} & 0.47{\scriptsize{$\pm$0.061}} & 0.63{\scriptsize{$\pm$0.026}} \\
DDIM+SA (ours) & & \textbf{0.99}{\scriptsize{$\pm$\textbf{0.006}}} & \textbf{0.99}{\scriptsize{$\pm$\textbf{0.004}}} & \textbf{0.96}{\scriptsize{$\pm$\textbf{0.014}}} & \textbf{0.99}{\scriptsize{$\pm$\textbf{0.010}}} & \textbf{0.93}{\scriptsize{$\pm$\textbf{0.026}}} & \textbf{0.87}{\scriptsize{$\pm$\textbf{0.025}}} & \textbf{0.57}{\scriptsize{$\pm$\textbf{0.052}}} & \textbf{0.78}{\scriptsize{$\pm$\textbf{0.045}}} \\
\midrule

3 step-CP & \multirow{2}{*}{3} & \textbf{0.99}{\scriptsize{$\pm$\textbf{0.007}}} & \textbf{0.99}{\scriptsize{$\pm$\textbf{0.007}}} & 0.93{\scriptsize{$\pm$0.052}} & \textbf{0.96}{\scriptsize{$\pm$\textbf{0.016}}} & 0.86{\scriptsize{$\pm$0.035}} & 0.83{\scriptsize{$\pm$0.039}} & 0.42{\scriptsize{$\pm$0.059}} & 0.49{\scriptsize{$\pm$0.051}} \\ 
3 step-CP+SA (ours) & & \textbf{0.99}{\scriptsize{$\pm$\textbf{0.004}}} & \textbf{0.99}{\scriptsize{$\pm$\textbf{0.005}}} & \textbf{0.94}{\scriptsize{$\pm$\textbf{0.056}}} & \textbf{0.96}{\scriptsize{$\pm$\textbf{0.020}}} & \textbf{0.93}{\scriptsize{$\pm$\textbf{0.023}}} & \textbf{0.89}{\scriptsize{$\pm$\textbf{0.032}}} & \textbf{0.53}{\scriptsize{$\pm$\textbf{0.077}}} & \textbf{0.62}{\scriptsize{$\pm$\textbf{0.057}}} \\
\midrule

1 step-CP & \multirow{2}{*}{1} & \textbf{0.99}{\scriptsize{$\pm$\textbf{0.007}}} & \textbf{0.99}{\scriptsize{$\pm$\textbf{0.006}}} & 0.87{\scriptsize{$\pm$0.073}} & 0.96{\scriptsize{$\pm$0.022}} & 0.87{\scriptsize{$\pm$0.025}} & 0.85{\scriptsize{$\pm$0.037}} & 0.27{\scriptsize{$\pm$0.071}} & 0.36{\scriptsize{$\pm$0.072}} \\ 
1 step-CP+SA (ours) & & \textbf{0.99}{\scriptsize{$\pm$\textbf{0.010}}} & \textbf{0.99}{\scriptsize{$\pm$\textbf{0.004}}} & \textbf{0.88}{\scriptsize{$\pm$\textbf{0.070}}} & \textbf{0.97}{\scriptsize{$\pm$\textbf{0.013}}} & \textbf{0.93}{\scriptsize{$\pm$\textbf{0.034}}} & \textbf{0.92}{\scriptsize{$\pm$\textbf{0.020}}} & \textbf{0.46}{\scriptsize{$\pm$\textbf{0.079}}} & \textbf{0.48}{\scriptsize{$\pm$\textbf{0.066}}} \\

\bottomrule
\end{tabular}
\label{tab:result_state_vision}
}
\end{table*}

\begin{figure}
    \centering
    \includegraphics[width=0.95\linewidth]{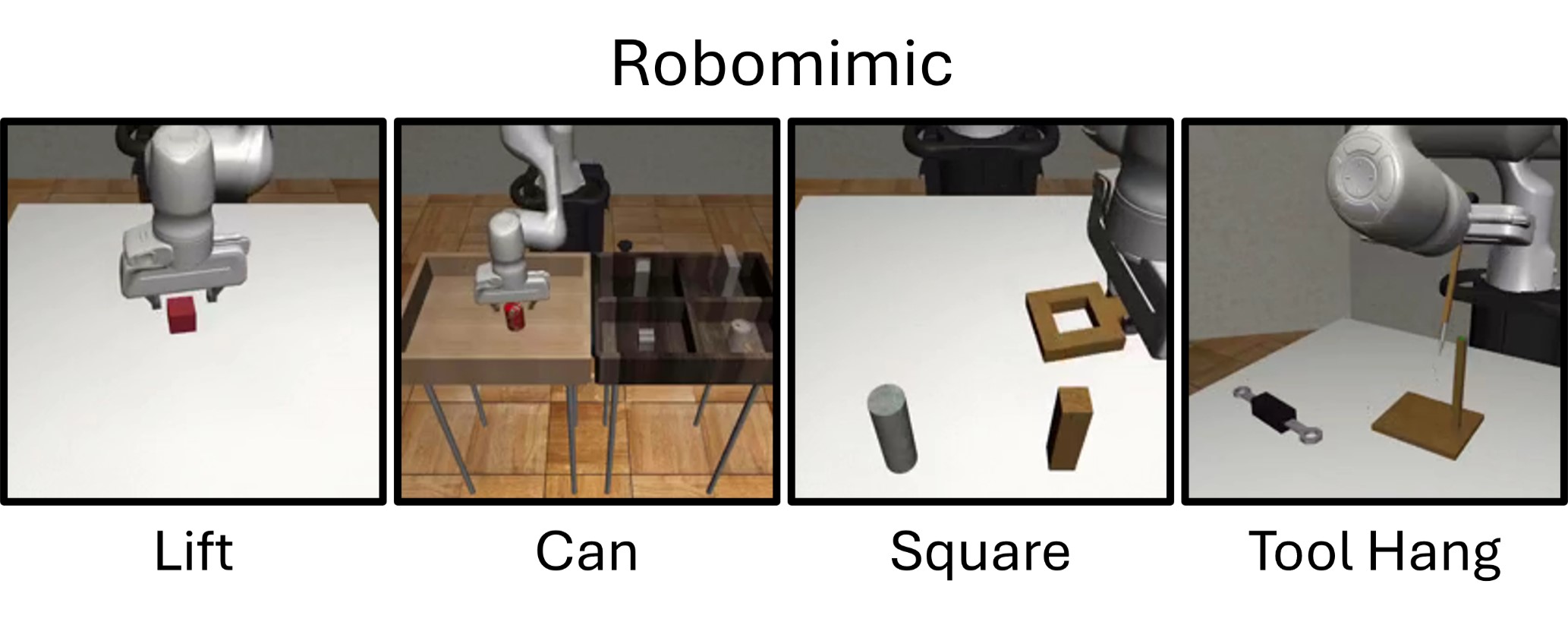}
    \vspace*{-3mm}
    \caption{\textbf{Robomimic Benchmark.}
    We evaluate our method against the baseline policies on single-robot manipulation tasks from the Robomimic~\cite{robomimic2021} benchmark.
    We categorize the Lift and Can tasks as ``Simple'' due to their lower precision requirements, while the Square and Tool Hang tasks are classified as ``Complex'' as they require precise manipulation.}
    \label{fig:task}
\end{figure}

\subsection{Standard Simulation Experiment} \label{subsec:simulation}

This section presents the evaluation results of the adaptive execution horizon with Spatial Attention in simulation, and compares its performance against the baseline policies.

\textbf{Tasks.} We evaluate our method and baselines on Robomimic~\cite{robomimic2021} benchmark, which has been used to evaluate Diffusion Policy~\cite{chi2023diffusion}, Consistency Policy~\cite{prasad2024consistency}, and ParaDiGMS~\cite{shih2023parallel}.
We evaluate our method and baselines on the Lift, Can, Square, and Tool Hang, all single-arm manipulation tasks.
Actions are defined as the 7-dimensional absolute end-effector position and orientation with respect to the robot base frame.
Demonstrations are collected via teleoperation by a proficient human operator, referred to as proficient human (PH) data in Robomimic~\cite{robomimic2021}.
Each task contains 200 successful demonstration episodes.
We classify Lift and Can as simple tasks, for which we set $T_{avg}$ to 16, and Square and Tool Hang as complex tasks, for which we set $T_{avg}$ to 32.
Figures of simulation tasks are provided in Figure~\ref{fig:task}.

\begin{figure}
    \centering
    \includegraphics[width=0.48\linewidth]{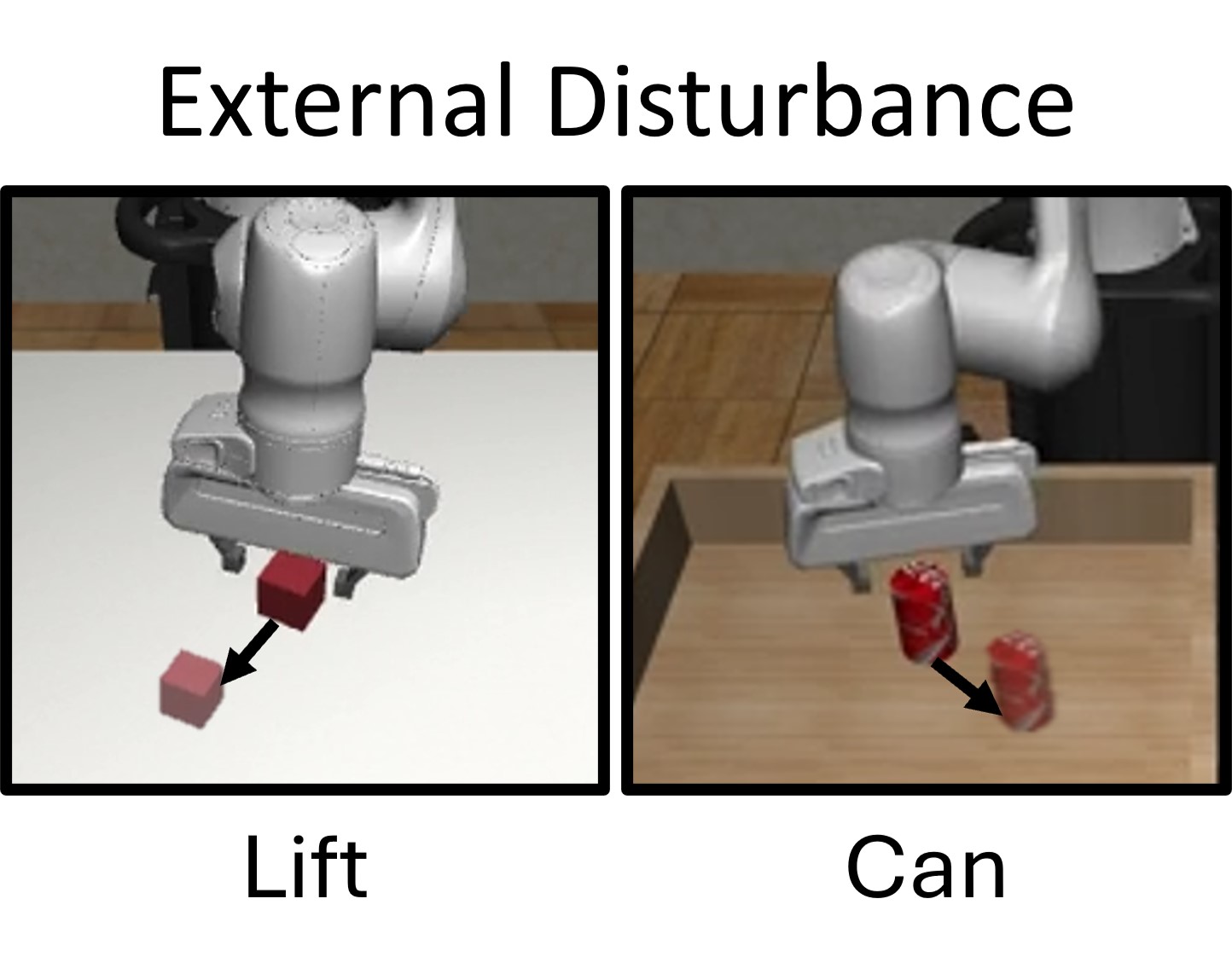}
    \vspace*{-3mm}
    \caption{\textbf{Lift/Can Task with Disturbance.}
    To evaluate the robustness of our algorithm against external disturbance, we modify the Lift (Left) and Can task (Right) task. These tasks are modified to move the target object away from the gripper once the gripper comes within a certain distance of it.}
    \label{fig:disturbance}
\end{figure}

\textbf{Results.}
Results of the policies under both state and image observation settings are presented in Table~\ref{tab:result_state_vision}.
The table reports $T_{avg}$ for each task.
Table~\ref{tab:result_state_vision} shows that using Spatial Attention to adaptively determine the execution horizon, denoted ``+SA'', consistently improves policy performance across all tasks.
It is important to note that all methods share identical policy architectures and parameters.

On the precision-critical tasks, policies equipped with Spatial Attention outperform the baselines on both state and vision observations by 5\%--7\% on Square and 10\%--19\% on Tool Hang.
Figure~\ref{fig:tool_hang} illustrates how Spatial Attention enables the policy to adaptively adjust $T_a$ in a way that results in a more responsive behavior while maintaining the same average execution horizon.
Spatial Attention takes high values in the phases that demand precision, such as grasping and inserting the target objects. 
Shortening $T_a$ in these phases lets the policy respond quickly to errors that arise during execution, improving the success rate without increasing the average computational cost.
\subsection{Responsiveness to External Disturbance}\label{subsec:responsiveness}

This section introduces a modified simulation environment in which external disturbances are applied to the target object.
In this environment, we evaluate the success rate of our method and the baseline algorithms to analyze their responsiveness to external disturbances.

\textbf{Task (External Disturbance).}
We customize the Lift and Can tasks from the Robomimic environment to build a simulation capable of applying disturbances.
In this experiment, disturbance is given by moving the target object.
As shown in Figure~\ref{fig:disturbance}, we move the target object in a direction orthogonal to the gripper once the gripper comes within a certain distance of the object.

\textbf{Results.}
Table ~\ref{tab:disturbance} reports the success rate of the baseline and our method under this experiment.
Since each algorithm uses the same $T_a$ and number of sampling steps as its counterpart in the previous section, we omit these values from the table.
According to Table~\ref{tab:disturbance}, applying Spatial Attention also outperforms all baselines in the disturbed environment.

\begin{figure}
    \centering
    \vspace{3mm}
    \includegraphics[width=0.95\linewidth]{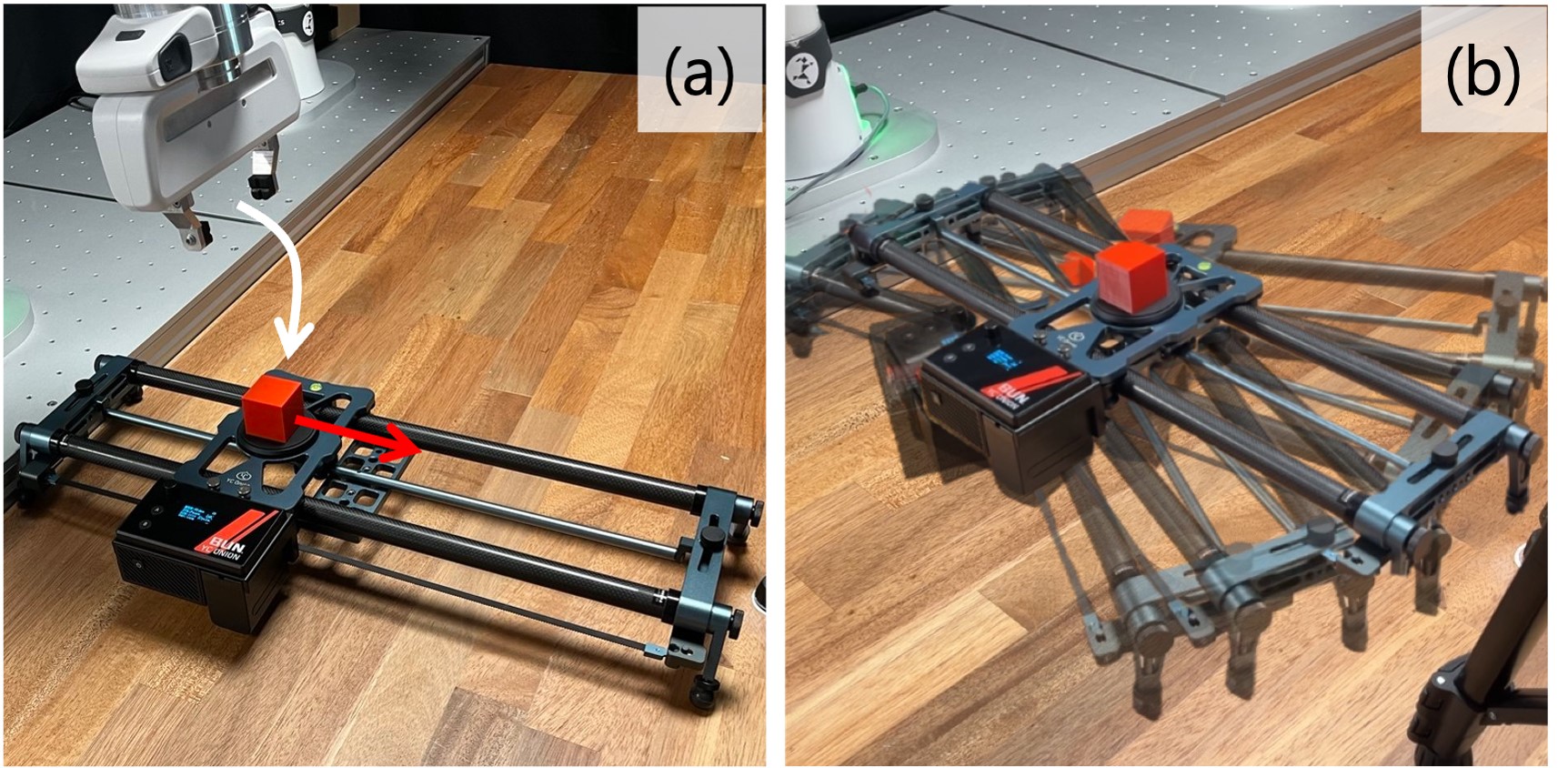}
    \vspace*{-3mm}
    \caption{\textbf{Real-World Experiment.} (a) Hardware setup. The robot must grasp the red cube (white arrow) while a camera slider moves the cube laterally (red arrow). (b) The slider is placed at 5 different orientations to vary the direction of the cube's motion relative to the robot. }
    \label{fig:realworld_setting}
\end{figure}

\subsection{Real-World Experiment}
In this section, we deploy a diffusion policy equipped with Spatial Attention on a real-world task and show that our framework can run in real time with improved robustness to disturbances.

\textbf{Task.}
The robot has to lift the red block moving on a camera slider with constant velocity shown in Figure~\ref{fig:realworld_setting}-(a).
We evaluate the policies by varying the camera slider direction across 5 orientations as shown in Figure~\ref{fig:realworld_setting}-(b).
For each orientation, a policy is evaluated 10 times, for a total of 50 rollouts.
In this experiment, DDIM is used as the backbone diffusion policy, since our objective is to demonstrate the advantage our algorithm provides for real-world deployment.

We use a 7-DoF Franka Research 3 as our manipulator and two Intel RealSense D435i cameras for vision observation, one mounted on the end-effector and one on the desk for the agent view.
For inference, we use a remote desktop powered by a 24GB RTX 4090 GPU.
TCP protocol sends observation data to remote desktop and sends sampled action chunk results to our robot platform.

We deploy a policy trained on the state-based observations of the Lift task.
To extract state observation from real-world vision observation, we utilize Foundation Pose~\cite{wen2024foundationpose}.
For comparison, vanilla DDIM is executed with a fixed execution horizon $T_a = 8$, while DDIM+SA is tuned to a comparable average execution horizon $\bar{T}_a \in [7.5, 8.5]$.

\begin{table}
\centering
\vspace{3mm}
\scriptsize
\setlength{\aboverulesep}{0pt}
\setlength{\belowrulesep}{0pt}
\renewcommand{\arraystretch}{1.2}
\caption{
\textbf{Evaluation results with external disturbance.} 
    We report the success rates of our method and its corresponding baselines, evaluated on the Robomimic Lift/Can tasks with disturbance. 
}
\begin{tabular}{@{}l|cc|cc@{}}
\toprule
\multirow{2}{*}{Method} & \multicolumn{2}{c|}{Lift} & \multicolumn{2}{c}{Can} \\ 
\cmidrule(lr){2-3} \cmidrule(lr){4-5}
 & State & Vision & State & Vision \\ 
 
\noalign{\global\aboverulesep=0.4ex \global\belowrulesep=0.65ex}
\midrule
DDPM & 0.58$\pm$0.040 & 0.65$\pm$0.032 & 0.76$\pm$0.040 & 0.74$\pm$0.022 \\
DDPM+SA (ours) & \textbf{0.67}$\pm$\textbf{0.034} & \textbf{0.75}$\pm$\textbf{0.051} & \textbf{0.84}$\pm$\textbf{0.049} & \textbf{0.78}$\pm$\textbf{0.029} \\
\midrule
DDIM & 0.53$\pm$0.055 & 0.59$\pm$0.052 & 0.77$\pm$0.032 & 0.76$\pm$0.025 \\
DDIM+SA (ours) & \textbf{0.60}$\pm$\textbf{0.057} & \textbf{0.73}$\pm$\textbf{0.063} & \textbf{0.86}$\pm$\textbf{0.022} & \textbf{0.79}$\pm$\textbf{0.031} \\
\midrule
3 step-CP & 0.60$\pm$0.071 & 0.66$\pm$0.053 & 0.68$\pm$0.050 & 0.70$\pm$0.038 \\
3 step-CP+SA (ours) & \textbf{0.63}$\pm$\textbf{0.060} & \textbf{0.80}$\pm$\textbf{0.044} & \textbf{0.73}$\pm$\textbf{0.050} & \textbf{0.72}$\pm$\textbf{0.047} \\
\midrule
1 step-CP & 0.64$\pm$0.092 & 0.56$\pm$0.103 & 0.63$\pm$0.074 & \textbf{0.68}$\pm$\textbf{0.041} \\
1 step-CP+SA (ours) & \textbf{0.67}$\pm$\textbf{0.085} & \textbf{0.72}$\pm$\textbf{0.102} & \textbf{0.68}$\pm$\textbf{0.078} & \textbf{0.68}$\pm$\textbf{0.048} \\
\bottomrule
\end{tabular}
\label{tab:disturbance}
\end{table}

\textbf{Results.}
Table~\ref{tab:realworld} reports the success rates of vanilla DDIM and DDIM with Spatial Attention (DDIM+SA). 
DDIM+SA substantially outperforms the baseline (0.92 vs.\ 0.42), consistent with the results in Sections~\ref{subsec:simulation} and~\ref{subsec:responsiveness}, even though DDIM+SA runs at a slightly longer average $T_a$ (8.3 vs.\ 8).
More specifically, DDIM fails to grasp the cube in most trials. 
As the gripper approaches the cube, its action chunk is too long to react to the cube's movement, resulting in a failed grasp.
DDIM+SA instead shortens the execution horizon once the gripper is near the cube, allowing it to adjust its motion to the cube's updated position. 

\begin{table}[th]
\centering
\caption{Real Robot experiment}
\begin{tabular}{@{}l|cc@{}}
\toprule
             & DDIM   & DDIM+SA (ours) \\ \midrule
Success Rate & 0.42  &  0.92           \\ 
Average $T_a$ & 8  &  8.3           \\ \bottomrule
\end{tabular}\label{tab:realworld}
\end{table}
\vspace{-2mm}

\section{Conclusion}
\label{sec:conclusion}

In this work, we introduce a framework that adaptively adjusts the execution horizon of action chunks for diffusion policies.
As the criterion for this adaptation, we define \textit{Spatial Attention}, a metric that quantifies the sensitivity of the policy's action distribution to variations in the observation input.
We show that adjusting the execution horizon to decrease as Spatial Attention increases minimizes the cumulative likelihood drop induced by disturbances under a fixed number of model inferences.
Experiments on the standard Robomimic benchmark and on perturbed scenarios show that adaptively adjusting the execution horizon with Spatial Attention outperforms fixed-length baselines, confirming that our framework achieves both computational efficiency and responsiveness.
Although we develop and evaluate our framework on diffusion policies in this work, it applies to any policy that generates action chunks.

A limitation, however, is that the framework requires a discrete action chunk.
Therefore it does not extend to streaming-based policies~\cite{chen2025responsive, hoeg2025fast, jiang2025streaming}, which generate actions continuously and expose no discrete chunk whose execution horizon can be adjusted; relaxing this restriction is an important direction for future work.
A further direction is to scale our framework to the large datasets used to train VLA models, a regime we have not yet explored.
The latent representation already learned by the VLM backbone could offer a natural space in which to evaluate Spatial Attention, making this a potentially practical avenue for the extension.

\bibliography{bibtex/bib/IEEEabrv.bib,bibtex/bib/IEEEexample.bib}{}
\bibliographystyle{IEEEtran}
\end{document}